\documentclass[11pt,a4paper]{article}
\usepackage[hyperref]{acl2020}
\usepackage{times}
\usepackage{latexsym}
\usepackage[TABBOTCAP]{subfigure}
\usepackage[shortlabels]{enumitem}

\usepackage{tikz-dependency}
\usepackage{algorithm}
\usepackage{algpseudocode}
\usepackage{multirow}
\usepackage{booktabs}

\usepackage{color}
\usepackage{helvet}
\usepackage{textcomp}
\usepackage{graphicx}
\graphicspath{ {images/} }
\usepackage{amsmath}
\usepackage{float}

\usepackage{hyperref}
\usepackage{url}

\aclfinalcopy 
\def\aclpaperid{*} 

\title{WinoWhy: A Deep Diagnosis of Essential Commonsense Knowledge \\for Answering Winograd Schema Challenge}

\author{Hongming Zhang\thanks{~~Equal contribution.} , Xinran Zhao$^{*}$, and Yangqiu Song \\
        Department of CSE, HKUST\\
        hzhangal@cse.ust.hk, xzhaoar@connect.ust.hk, yqsong@cse.ust.hk}
\date{}

\begin{document}
\maketitle
\begin{abstract}
In this paper, we present the first comprehensive categorization of essential commonsense knowledge for answering the Winograd Schema Challenge (WSC).
For each of the questions, we invite annotators to first provide reasons for making correct decisions and then categorize them into six major knowledge categories.
By doing so, we better understand the limitation of existing methods (i.e., what kind of knowledge cannot be effectively represented or inferred with existing methods) and shed some light on the commonsense knowledge that we need to acquire in the future for better commonsense reasoning.
Moreover, to investigate whether current WSC models can understand the commonsense or they simply solve the WSC questions based on the statistical bias of the dataset, we leverage the collected reasons to develop a new task called WinoWhy, which requires models to distinguish plausible reasons from very similar but wrong reasons for all WSC questions.
Experimental results prove that even though pre-trained language representation models have achieved promising progress on the original WSC dataset, they are still struggling at WinoWhy.
Further experiments show that even though supervised models can achieve better performance, the performance of these models can be sensitive to the dataset distribution.
WinoWhy and all codes are available at: \url{https://github.com/HKUST-KnowComp/WinoWhy}.

\end{abstract}

\section{Introduction}\label{sec-introduction}

\begin{figure}[t]
    \centering
    \includegraphics[width=0.8\linewidth]{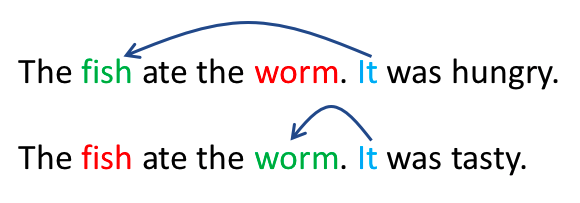}
    \caption{A pair of questions in WSC.}
    \label{fig:wino_example}
\end{figure}

Commonsense reasoning, as an important problem of natural language understanding, has attracted much more  attention in the NLP community recently~\cite{levesque2012winograd,zhou2018commonsense,ostermann2018semeval,talmor2019commonsenseqa}. 
Among all developed commonsense reasoning tasks, the Winograd Schema Challenge (WSC)~\cite{levesque2012winograd}, which is a hard pronoun coreference resolution task, is one of the most influential ones.
All questions in WSC are grouped into pairs such that paired questions have minor differences (mostly one-word difference), but reversed answers. 
For each question, we denote the other question in the same pair as its reverse question. 
One pair of the WSC task is shown in Figure~\ref{fig:wino_example}.
Based on the design guideline of WSC, all commonly used features (e.g., gender, plurality, and co-occurrence frequency) do not have any effect.
Human beings can solve these questions because of their shared commonsense knowledge.
For example, ordinary people can know that the pronoun `it' in the first sentence refers to `fish' while the one in the second sentence refers to `worm' because `hungry' is a common property of something eating things while `tasty' is a common property of something being eaten.

Conventionally, people tried to leverage crowd-sourced commonsense knowledge bases~\cite{DBLP:conf/aaaiss/LiuJLZWH17} or search engines~\cite{DBLP:conf/emnlp/EmamiCTSC18} to solve the WSC task, but performances of these models are not satisfying.
Recently, pre-trained language representation models~\cite{DBLP:conf/acl/KocijanCCYL19,radford2019language,liu2019roberta} have demonstrated significant improvements in both unsupervised and supervised settings.
However, as these approaches treat the concept `commonsense knowledge' as a black box, we are not clear about why they can do better (e.g., can these models understand commonsense or they just capture the statistical bias of the dataset) and do not know how to further improve them.
To answer these two questions, in this work, we present the first deep diagnosis of essential commonsense knowledge for answering WSC questions.
Specifically, we invite annotators to first provide reasons for why they choose the answers when they answer the questions, and then group all the WSC questions by different types of used commonsense knowledge (e.g., the property of entities, temporal knowledge, or spatial knowledge).
By doing so, we can then analyze what kinds of commonsense knowledge can be well represented and understood by current models and more importantly, we can be clear about what kinds of commonsense knowledge are still challenging for current models, which could be an important future research direction for solving not only the WSC task but also the general commonsense reasoning problem.

\begin{figure}[t]
    \centering
    \includegraphics[width=\linewidth]{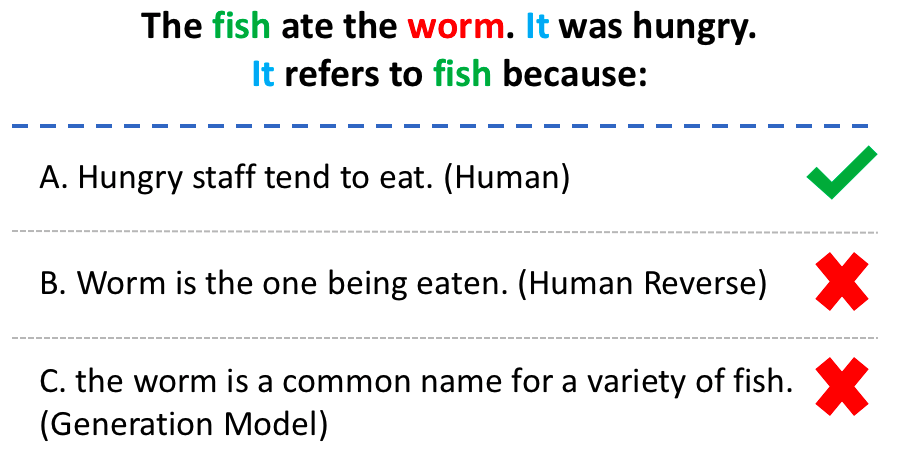}
    \caption{One example from the WinoWhy dataset. Plausible and implausible reasons are indicated with the tick and the crosses respectively. Resources of different reasons are shown in brackets. `Human Reverse' means the human reason for the reverse question.}
    \label{fig:winowhy_example}
\end{figure}

After the diagnosis, based on the collected reasons, we also create a new task WinoWhy, which aims at better evaluating models' abilities to understand commonsense knowledge. 
For each question in the WSC task, we pair it with several reasons. Models are required to distinguish the correct reasons from all very similar but wrong candidates.
From examples in Figure~\ref{fig:winowhy_example}, we can see that even though all candidates are highly related to the original question, only one of them is the correct reason for resolving the coreference relation.
Experimental results show that even though state-of-the-art models can achieve about 90\% accuracy on the original WSC task, they are still struggling on WinoWhy questions, which shows that current models are still far away from understanding the commonsense knowledge.
Moreover, by conducting experiments on both WSC and WinoWhy tasks, we prove that even though supervised models can achieve better performance, these models can be sensitive to the dataset distribution, which indicates that the improvement is probably coming from better capturing the statistical bias of the dataset rather than better understanding the required commonsense knowledge.

The rest of the paper is organized as follows. 
In Section~\ref{sec:diagnosis}, we present the diagnosis of essential commonsense knowledge for answering WSC questions, which includes the reason collection and categorization.
After that, we show how we create WinoWhy in Section~\ref{sec:winowhy}.
In Sections~\ref{sec:wsc-exp} and~\ref{sec:winowhy-exp}, we introduce the detailed experiments and analysis on both the original WSC and the proposed WinoWhy tasks.
We introduce the related work about commonsense reasoning in Section~\ref{sec:related-work}. 
In the end, we conclude this paper with Section~\ref{sec:conclusion}.

\section{Commonsense Knowledge Diagnosis}\label{sec:diagnosis}

Commonsense reasoning is often viewed as one of the most challenging AI tasks and we still do not have a principled way of solving it. 
One important reason behind this is that, due to the vague definition of commonsense knowledge, we are not clear about what the essential knowledge types are and thus we are unclear about how to represent, acquire, and use them.
As a result, we can only treat commonsense knowledge as a black box and try to learn it from limited training data.
To explore a principled way of representing commonsense knowledge and solving commonsense reasoning problems, we take the Winograd Schema Challenge as the breaking point to conduct a detailed diagnosis of what kinds of knowledge are essential for answering these questions.
To be specific, we first ask human beings to provide reasons why they make the correct decisions for all WSC questions.
After that, we categorize these reasons by the involved knowledge types (e.g., the property of objects, temporal knowledge, or spatial knowledge).
By doing so, we are more clear about how to acquire, represent, and apply such knowledge.
Details are introduced as follows.

\subsection{Reason Collection}\label{sec:reason-collection}

\begin{figure}[t]
    \centering
    \includegraphics[width=\linewidth]{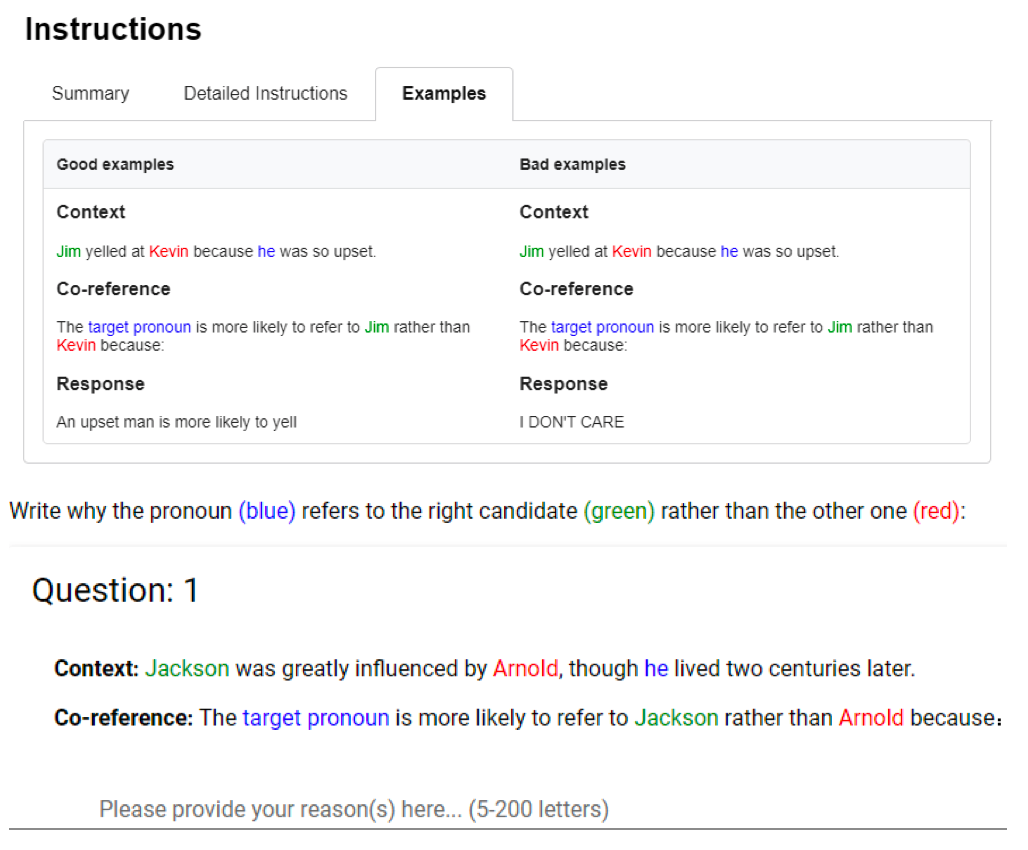}
    \caption{Reason collection interface on MTurk. Annotators are asked to provide reason(s) for all WSC questions in natural language.}
    \label{fig:survey_example_1}
\end{figure}

To collect high-quality reasons for answering all WSC questions, we employ the Amazon Mechanical Turk  (MTurk) platform
for our annotations and design a two-phase annotation procedure to collect the knowledge. 
In the first phase, we ask annotators to provide reasons for all WSC questions. Detailed instructions are provided such that annotators can fully understand the task\footnote{Instructions are shown in Appendix ~\ref{appendix:reason_collection}.}.
As each question may have multiple plausible reasons, for each question, we invite five annotators to provide reasons based on their own judgments.
A screenshot of the survey is shown in Figure~\ref{fig:survey_example_1}.
As a result, we collect 1,365 reasons. 
As the quality of some given reasons might not be satisfying, we introduce the second round annotation to evaluate the quality of collected reasons.
In the second phase, for each reason, we invite five annotators to verify whether they think the reason is reasonable or not\footnote{Survey examples are shown in Appendix ~\ref{appendix:reason_plausibility}.}. If at least four annotators think the reason is plausible, we will accept that reason.
As a result, we identify 992 valid reasons.




\subsection{Knowledge Categorization}

After collecting all reasons, we categorize them into different groups based on the used knowledge types. 
We first introduce the selected knowledge types and then introduce the detailed annotation procedure.

\subsubsection{Knowledge Types}

A good categorization standard should have two properties: (1) Broad Coverage: it should cover most cases; (2) Exclusive: there should be clear boundaries between different categories. Following these standards, we found following two categorization methods of commonsense knowledge:
\begin{enumerate}[leftmargin=*]
    \item \textbf{Conceptual Semantic Theory}: According to Jackendoff's original theory~\cite{Jackendoff}, the semantics of human language can be expressed with a finite set of mental primitives and a finite set of principles of mental combination. As claimed by Jackendoff, even though the definition of mental primitives may vary based on different data or languages, some common primitives (i.e., \textit{entity}, \textit{property}, \textit{number}, \textit{location}, \textit{state}, \textit{event}, and \textit{activity}) can be observed. These common primitives can thus be used as knowledge types for the commonsense knowledge categorization.
    \item \textbf{ConceptNet}: As one of the most popular commonsense knowledge resources, ConceptNet 1.0~\cite{liu2004conceptnet} defines 20 commonsense relations, which belong to eight categories (i.e., \textit{K-lines}, \textit{Things}, \textit{Agents}, \textit{Events}, \textit{Spatial}, \textit{Causal}, \textit{Functional}, and \textit{Affective}). In the latest version of ConceptNet~\cite{DBLP:conf/aaai/SpeerCH17}, more relations (e.g., `RelatedTo') from other resources are merged into ConceptNet. As they are relatively vague, we still follow the definition in ConceptNet 1.0 for the commonsense knowledge categorization.
\end{enumerate}

As there exist some overlaps between semantic primitives and categories in ConceptNet (e.g., `\textit{Agents}' and `\textit{Functional}' both describe certain properties of some objects), we first adopt all the commonly observed primitives in~\cite{Jackendoff} as the base knowledge types and then modify them based on the definition of categories from ConceptNet. 
For example, three primitives (\textit{activity}, \textit{state}, and \textit{event}) and \textit{Events} from ConceptNet can all be covered by the definition of Eventuality~\cite{ALEXANDER1978}. For the simplicity of the categorization and the quality of the annotation, we merge them.
At the current stage, we remove `\textit{K-lines}' because it contains relations like `ConceptuallyRelatedTo', which is relatively vague and difficult to be distinguished from other categories.
Another exceptional knowledge type is `\textit{Causal}' from ConceptNet.
During the annotation, we found out that annotators had difficulty understanding the strict definition of `\textit{Causal}' in ConceptNet (i.e., One event contributes to the creation of another one) and tended to annotate all reasons as `\textit{Causal}' because they think all reasons can somehow `cause' the decision making.
To make sure that all categories are easy for annotators, which are mostly ordinary people, to distinguish, we remove `Causal'.
As we cannot guarantee that selected knowledge types could cover all kinds of knowledge, an additional type `Others' is provided.
Names, definitions, and examples of selected knowledge types are shown in Table~\ref{tab:knowledge_type}.

\begin{table}[t]
\small
    \centering
    \begin{tabular}{p{1.3cm}|p{2.7cm}|p{2.4cm}}
    \toprule
    Name     & Definition & Example\\
         \midrule
       Property  & Knowledge about property of objects. & ice is cold.\\
       \hline
       Object    & Knowledge about objects. & cats have ears. \\
       \hline
       Eventuality  &  Knowledge about eventualities. & `wake up' happens before `open eyes'.\\
       \hline
       Spatial   & Knowledge about spatial position. & object at the back can be blocked.\\
       \hline
       Quantity  & Knowledge about numbers. & 2 is smaller than 10.\\
       \hline
       Others    & All other knowledge. & NA \\
    \bottomrule
    \end{tabular}
    \caption{Names, definitions, and examples of selected knowledge types. Annotators are asked to select the most suitable knowledge type of each reason. If they think none of the first five categories is suitable, they are encouraged to choose `Others'.}
    \label{tab:knowledge_type}
\end{table}

\subsubsection{Annotation}

\begin{figure}[t]
    \centering
    \includegraphics[width=\linewidth]{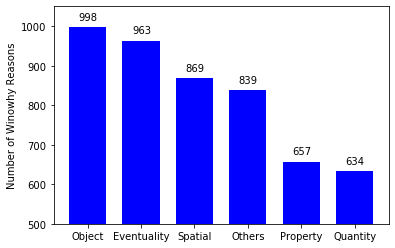}
    \caption{Distribution of different knowledge types.}
    \label{fig:reason-category}
\end{figure}

For each collected valid reason, we invite annotators to select the knowledge type that can best describe the reason\footnote{Survey examples are shown in Appendix ~\ref{appendix:reason_categorization}.}. Note that each reason may contain inference over multiple knowledge types.
Thus, for each reason, we invite five different annotators to provide annotations.
Each annotators are provided with detailed instruction of the job, descriptions of each candidate category, and examples for the category. As a result, we collect 4,960 annotations.
We show the distribution of annotation results in Figure~\ref{fig:reason-category}.
From the distribution, we can see that all knowledge types are very important, especially the knowledge about objects (e.g., `cats have ears') and eventualities (e.g., `people who give help often receive thanks later').
Besides that, we also notice that only 17\% of all reason annotations (839) are `Others', which indicates that the selected five categories can effectively cover 83\% of the cases and thus the selected knowledge types fulfill the broad coverage requirement.
We evaluate the annotation quality by average inner annotator agreement (IAA) and kappa coefficient~\cite{kappa}. We compute the IAA pair-wisely among all annotators. For each reason, if two annotators give the same knowledge type, we label it as agreed, otherwise, we label it as dis-agreed.
The average IAA is 78.72\%. We calculate the kappa coefficient based on the five raters and five categories setting and the result is 0.804. Considering that the annotation task is a multiple-choice task, such an agreement can indicate that the survey is well designed and annotators can clearly understand the task.
For each WSC question, we select the most popular knowledge type among all valid reasons as the question's major knowledge type. If multiple knowledge types have the same votes, we assume that question has multiple knowledge types. As a result, 222 questions have single knowledge type and 51 questions have multiple knowledge types.

\section{WinoWhy}\label{sec:winowhy}

In this section, we introduce details about the creation of WinoWhy.

\subsection{Task Definition}

Each question in WinoWhy is defined as follows. Given a pronoun coreference resolution question and its correct answer from the original WSC data, models are asked to select all plausible reasons for making the correct prediction. 
WinoWhy can thus be viewed as a natural followup of the original WSC task and can help better understand models' commonsense reasoning abilities.

\subsection{Candidate Selection}

For each question, three kinds of candidate reasons are selected for annotators to annotate. 
The first reason resource is human annotation, which effectively represents how human beings solve these questions.
Besides that, to collect some very similar but wrong reasons as negative examples, we consider the reasons provided by humans for the reverse question as a potential challenging wrong reason resource.
Last but not least, besides reasons provided by human beings, we also leverage a strong generation model (i.e., GPT-2~\cite{radford2019language}) to generate reasons.
We provide the same questions that we showed to humans before (e.g., `The fish are the worm. it was hungry. It refers to fish because') to the generation model and ask it to finish the sentences.
For each question, we leverage the beam search to find the top five generated reasons.
Merging all resources, we get 4,095 reasons for the next step annotation.

%

\subsection{Annotations}
\begin{figure}[t]
    \centering
    \includegraphics[width=\linewidth]{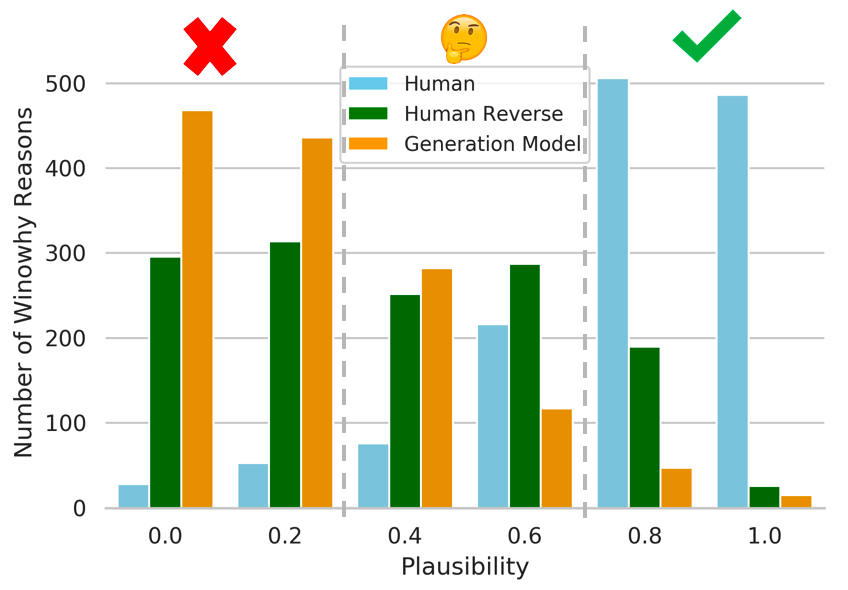}
    \caption{Distribution of reason plausibility score. The positive, acceptable, and negative reasons are denoted with the tick, confusing emoji, and cross respectively.}
    \label{fig:reason-quality}
\end{figure}

Similar to previous annotations, we invite annotators from Amazon Turk to help annotate whether the reasons are plausible or not.
For each reason, we invite five different annotators and determine the plausibility score of each reason by voting. For example, if four out of the five annotators think one reason is plausible, its plausibility score is then 0.8.
We use the same survey to annotate the plausibility of different reasons as Section~\ref{sec:reason-collection}.
As a result, we collect 20,475 annotations. The average IAA is 91.49\% and the kappa coefficient (five raters and two categories) is 0.880.

\subsection{Dataset Analysis}



We show the distribution of annotation results in Figure~\ref{fig:reason-quality}, from which we can make the following observations.
First, most of the reasons given by humans are reasonable, which fits our previous observation.
Second, even though the majority of reverse reasons are not plausible, which fits our assumption, some of them do make sense.
One scenario is that when the reason is comparing some property of both candidates, it can be used for both questions.
For example, for the question pair ``The trophy doesn't fit into the brown suitcase because it is too small/large'', explanations like ``Only small objects can fit into large objects'' are plausible for both questions.
Last but not least, not surprisingly, most of the reasons generated by GPT-2 have relatively low quality.
To analyze why the reasons generated by GPT-2 are not satisfying, we show one example in Table~\ref{tab:gpt-example}.
Based on the five reasons, we can find two limitations of GPT-2: (1) it could generate some meaningless words (e.g., `-C.B.'), which could influence the overall quality significantly; (2) some of the answers are related and complete sentences by themselves, but they are not a valid reason for the question. For example, the second reason is wrong because Charlie cannot be the one who has given the money. 
These observations show that understanding commonsense knowledge is still a challenging task for current pre-trained language representation models like GPT-2.

If at least four out of five annotators regard one reason as plausible, we label it as a positive reason. If only one or zero annotators think it is plausible, we label it as a negative reason. All others are labeled as acceptable reasons.
To ensure the clear boundary between positive and negative examples in WinoWhy, only positive and negative reasons are selected to evaluate models.
In total, WinoWhy contains 1,270 positive and 1,595 negative examples.

\begin{table}[t]
\small
    \centering
    \begin{tabular}{p{5cm}|p{1.5cm}}
    \toprule
        Reason & Plausibility Score \\
         \midrule
         of the circumstances of his birth." -C.B. & 0.0/1.0 \\
        \hline
         he's the one who's given him the money to do so. & 0.2/1.0 \\
        \hline
         it was Charlie who started the discussion and who encouraged Charlie to take up the challenge. & 0.0/1.0\\
        \hline
        we feel grateful for the help from others & 1.0/1.0 \\
        \hline
        charlie is the one who get help. & 0.6/1.0 \\
         \bottomrule
    \end{tabular}
    \caption{Given the sentence ``Bob paid for Charlie's college education. He is very grateful. The `He' refers to Charlie because '', the reasons generated by GPT-2 and corresponding plausibility scores.}
    \label{tab:gpt-example}
\end{table}

\section{WSC Experiments}\label{sec:wsc-exp}
In this section, we present the performance of current models on WSC. By doing so, we can better understand their strengths and limitations.


\subsection{Evaluated Methods and Implementation}

Recently, pre-trained language representation models have achieved significant improvement on the WSC task. In this section, we evaluate the following three models:
\begin{enumerate}[leftmargin=*]
    \item \textbf{BERT}~\cite{DBLP:conf/naacl/DevlinCLT19}: As a powerful contextualized word representation model, it has been proven helpful in many downstream NLP tasks.
    As shown in~\cite{DBLP:conf/acl/KocijanCCYL19}, we can first convert the original WSC task into a token prediction task and then leverage BERT to solve the problem. We denote the base and large model of BERT as BERT (base) and BERT (large) respectively.
    \item \textbf{GPT-2}~\cite{radford2019language}: GPT-2 is one of the best pre-trained language models for generation tasks. As reported in the original paper, we can first replace the pronouns with different candidates and leverage the probability of the {\it full} or {\it partial} sentences to make the prediction. Here we evaluate the small (117 M parameters) and the large (774 M parameters) models and denote those settings as GPT-2 (small, full), GPT-2 (small, partial), GPT-2 (large, full), and GPT-2 (large, partial) respectively. 
    \item \textbf{RoBERTa}~\cite{liu2019roberta}: RoBERTa is a recent improved version of BERT with larger amount of training instances and techniques such as dynamic masking, which performs consistently better than BERT over many benchmark datasets. We denote the base and large models of RoBERTa as RoBERTa (base) and RoBERTa (large) respectively. 
\end{enumerate}
Besides unsupervised models, as indicated by~\cite{DBLP:conf/acl/KocijanCCYL19}, fine-tuning BERT with a similar pronoun resolution dataset WSCR~\cite{DBLP:conf/emnlp/RahmanN12} can help boost the performance. A later work \cite{sakaguchi2019winogrande} has further enhanced the performance by fine-tuning RoBERTa with a larger and more balanced dataset WinoGrande. Statistics of these datasets are presented in Table~\ref{tab:dataset-stats}. In our experiments, we evaluate the combination of different pre-trained models and fine-tuning datasets, and denote them as BERT (base/large) + WSCR/Grande and RoBERTa (base/large) + WSCR/Grande respectively.

\begin{table}[t]
\small
    \centering
    \begin{tabular}{p{1.6cm}|p{1.3cm}|p{1.3cm}|p{1.3cm}}
    \toprule
        Dataset & \#Problems & Average Length & \#Vocab \\
         \midrule
        WSC & 273 & 19.1 & 919 \\
        WSCR & 1,886 & 15.9 & 4,127 \\
        WinoGrande & 43,972 & 20.6 & 16,469\\
         \bottomrule
    \end{tabular}
    \caption{Statistics of WSC and related datasets.}
    \label{tab:dataset-stats}
\end{table}

\begin{table*}[t]
\small
    \centering
    \begin{tabular}{l||c|c|c|c|c|c||c}
    \toprule
       Model  & Property & Object & Eventuality & Spatial & Quantity & Others & Overall\\
         & (32) & (82) & (88) & (64) & (20) & (48) & (273) \\
    \midrule
        BERT (base) &56.25\% &64.63\% &50.00\% &57.81\% &50.00\% &45.83\% &56.04\%\\
        BERT (large) &56.25\% &62.20\% &62.50\% &67.19\% &45.00\% &52.08\% &61.90\%\\
        RoBERTa (base) &43.75\% &51.22\% &56.82\% &51.56\% &55.00\% &39.58\% &51.65\%\\
        RoBERTa (large) &50.00\% &51.22\% &52.27\% &48.44\% &65.00\% &56.25\% &52.75\%\\
        
         \midrule
        GPT-2 (small, full) &56.25\% &51.22\% &55.68\% &51.56\% &60.00\% &47.92\% &52.75\%\\
        GPT-2 (small, partial) &43.75\% &60.98\% &53.41\% &51.56\% &60.00\% &54.17\% &53.48\%\\
        GPT-2 (large, full) &68.75\% & 68.29\% &61.36\% &53.13\% &55.00\% &45.83\% &59.34\%\\
        GPT-2 (large, partial) &65.63\% &75.61\% &72.73\% &62.50\% &65.00\% &60.42\% &69.23\%\\
        \midrule
        BERT (base) + WSCR &71.88\% &64.63\% &55.68\% &59.38\% &65.00\% &45.83\% &59.71\%\\
        BERT (large) + WSCR &81.25\% &75.61\% &73.86\% &67.19\% &85.00\% &64.58\% &71.43\%\\
        BERT (base) + Grande &65.63\% &58.54\% &60.23\% &59.38\% &55.00\% &56.25\% &60.34\% \\
        BERT (large) + Grande &75.00\% &70.73\% &77.27\% &79.69\% &75.00\% &68.75\% &73.63\% \\
        \midrule
        RoBERTa (base) + WSCR &62.50\% &60.98\% &57.95\% &64.06\% &55.00\% &64.58\% &63.00\% \\
        RoBERTa (large) + WSCR  &84.38\% &\textbf{84.15\%} &79.55\% &76.56\% &70.00\% &81.25\% &80.95\% \\
        RoBERTa (base) + Grande &75.00\% &67.07\% &72.73\% &75.00\% &80.00\% &70.83\% &72.16\% \\
        RoBERTa (large) + Grande &\textbf{90.63\%} &\textbf{84.15\%} &\textbf{93.18\%} &\textbf{84.38\%} &\textbf{90.00\%} &\textbf{89.58\%} &\textbf{87.55\%}\\
    \bottomrule
    \end{tabular}
    \caption{Performances of different models on WSC questions. Questions are grouped by their major knowledge types. If one question contains more than one knowledge types, it will be counted in all categories. If one question contains only `Others' knowledge, it will be grouped into `Others'. Numbers of questions are shown in brackets.}
    \label{tab:wsc-main-result}
\end{table*}




\begin{table}[t]
\small
    \centering
    \begin{tabular}{l||c|c}
    \toprule
        Model & Single & Multiple\\
        & (222) & (51)\\
    \midrule
        BERT (base) &56.31\% &54.90\% \\
        BERT (large) &63.06\% &56.86\% \\
        RoBERTa (base) &53.15\% &45.10\% \\
        RoBERTa (large) &54.05\% &47.06\% \\
        \midrule
        GPT-2 (small, full) &51.80\% &56.86\% \\
        GPT-2 (small, partial) &53.48\% &54.90\% \\
        GPT-2 (large, full) &58.56\% &62.74\% \\
        GPT-2 (large, partial) &70.27\% &64.71\% \\
        \midrule
        BERT (base) + WSCR &59.91\% &58.82\% \\
        BERT (large) + WSCR &70.27\% &76.47\% \\
        BERT (base) + Grande &61.26\% &56.86\% \\
        BERT (large) + Grande &72.52\% &78.43\% \\
         \midrule
        RoBERTa (base) + WSCR &64.86\% &54.90\% \\
        RoBERTa (large) + WSCR &81.53\% &78.43\% \\
        RoBERTa (base) + Grande &72.97\% &68.63\% \\
        RoBERTa (large) + Grande &\textbf{86.94\%} &\textbf{90.20\%} \\
         
    \bottomrule
    \end{tabular}
    \caption{Performances of different models on different sets of WSC questions. Questions are grouped by the number of essential knowledge types (i.e., single or multiple). Numbers of questions are shown in brackets.}
    \label{tab:wsc-result-type-number}
    \vspace{-0.2in}
\end{table}

\begin{table*}[t]
\small
    \centering
    \begin{tabular}{l||c|c|c|c|c|c||c}
    \toprule
        Model  & Property & Object & Eventuality & Spatial & Quantity & Others & Overall\\
         & (337) & (856) & (928) & (674) & (206) & (496) & (2865) \\
        
    \midrule
         Majority Voting &54.30\% &56.31\% &56.47\% &52.67\% &52.43\% &55.24\% &55.67\% \\
         \midrule
         BERT (base) &56.97\%  &56.54\% &56.25\%  &54.01\%  &51.94\%  &55.44\%  &55.92\%  \\
         BERT (large) &56.38\% &57.24\% &56.14\% &53.41\% &51.94\% &56.65\% &56.13\% \\
         RoBERTa (base)  &54.30\% &56.31\% &56.90\% &52.67\% &52.91\% &55.44\% &55.78\%\\
         RoBERTa (large) &54.30\% &56.43\% &56.47\% &52.67\% &52.43\% &55.04\% &55.67\%\\
         GPT-2 (small) &56.68\%  &54.91\% &57.11\%  &54.45\%  &\textbf{59.71\%}  &57.66\%  &56.37\%  \\
         GPT-2 (large) &\textbf{57.57\%} &54.44\%&54.42\% &55.93\% &54.85\% &54.84\% &55.77\% \\
         \midrule
         BERT (base) + WSCR &55.49\% &56.31\% &56.90\% &52.97\% &51.94\% &55.04\% &55.71\% \\
         BERT (large) + WSCR &56.97\% &56.31\% &56.79\% &53.12\% &52.91\% &55.04\% &55.99\% \\
         BERT (base) + Grande &57.27\% &56.43\% &57.22\% &53.41\% &52.91\% &55.24\% &55.99\% \\
         BERT (large) + Grande &54.90\% &56.07\% &56.57\% &52.67\% &52.91\% &55.44\% &55.71\% \\
        \midrule
        RoBERTa (base) + WSCR &52.82\% &55.61\% &58.41\% &53.26\% &56.31\% &55.04\% &56.19\% \\
        RoBERTa (large) + WSCR &54.90\% &58.06\% &56.90\% &52.08\% &52.91\% &56.85\% &56.23\% \\
        RoBERTa (base) + Grande &56.08\% &\textbf{58.88\%} &58.19\% &55.64\% &57.28\% &57.66\% &58.05\% \\
        RoBERTa (large) + Grande &56.08\% &58.06\% &\textbf{59.59\%} &\textbf{56.82\%} &56.80\% &\textbf{58.06\%} &\textbf{58.18\%} \\
    \bottomrule
    \end{tabular}
    \caption{Performances of different models on WinoWhy questions. 
    We report performances of different reason sets based on the required knowledge types. Reasons could belong to multiple categories as the original WSC questions could contain more than one knowledge types. Numbers of questions are shown in brackets.}
    \label{tab:winowhy-main-result}
\end{table*}

\subsection{Result Analysis}

From the result in Table~\ref{tab:wsc-main-result}, we can make following observations: 
(1) Larger models perform better on all knowledge types due to their stronger semantic representation abilities; 
(2) The partial version of GPT-2 significantly outperforms the full version, which is consistent with the observation in~\cite{DBLP:journals/corr/abs-1806-02847} and is mainly because the influence of imbalanced distribution of candidate words are relieved by only considering the sentence probability after the pronouns. Such observation also explains why GPT-2 can outperform unsupervised BERT on WSC because models based on BERT, which rely on predicting the probability of candidate words, cannot get rid of such noise; 
(3) For most models, questions that require spatial knowledge are the most challenging ones. One possible explanation is that the inference over spatial knowledge is often triggered by a preposition (e.g., `in' or `behind'), which is challenging for language representation models to remember without enough training corpus for spatial knowledge specifically; 
(4) Questions belonging to `Others' involve more complex inference, even over multiple types of knowledge and thus most models perform poorly on that. The only exception is RoBERTa, which leverages its strong language representation ability to overcome such a challenge; 
(5) Fine-tuning over WinoGrande significantly boosts the performance.

Besides the above analysis, we are also interested in how different models perform on questions that require complex reasoning types. Thus we divide all WSC questions based on how many knowledge types are required to solve these questions and show the result in Table~\ref{tab:wsc-result-type-number}.
Based on the result, we can see that relatively small models (e.g., BERT (base) and RoBERTa (base)) perform better on questions that require single knowledge types rather than multiple knowledge types.
However, for large models (e.g., BERT (large) and RoBERTa (large)), as long as the suitable fine-tuning dataset is provided, they can achieve similar and even better performance on the complicated questions. In general, this observation is consistent with our previous observations that large models are capable of solving complex questions from the `Others' category with the support of suitable fine-tuning datasets.

\section{WinoWhy Experiments}\label{sec:winowhy-exp}

In this section, we conduct experiments to investigate whether current models can understand how human beings solve WSC questions.

\subsection{Unsupervised Setting}

\noindent \textbf{Experiment Details:} To evaluate whether pre-trained language representation models, which achieve the state-of-the-art performance on the WSC task, can distinguish the plausible reasons against the wrong ones, following~\cite{DBLP:conf/acl/KocijanCCYL19,radford2019language,sakaguchi2019winogrande}, we first connect the questions and candidate reasons into single sentences, put them into the models, and take the returned probability as the prediction. Higher probability indicates higher plausibility prediction. Best thresholds are selected for different models to calculate the final accuracy. Similar to Section~\ref{sec:wsc-exp}, we evaluate BERT (base), BERT (large), GPT-2 (small), GPT-2 (large), RoBERTa (base), and RoBERTa (large) on WinoWhy.
For GPT-2 models, as the partial setting has been proved more useful, we only report the performances based on the partial setting. 
Besides these two, we also consider BERT/RoBERTa + WSCR/Grande combinations as additional unsupervised approaches because they are not directly optimized towards the WinoWhy task.

\noindent \textbf{Result Analysis:} Based on the results shown in Table~\ref{tab:winowhy-main-result}, we can observe that even though pre-trained language representation models have achieved significant improvement over the original WSC task, they are still struggling on the WinoWhy task. Moreover, experimental results on different knowledge types prove that such a conclusion is universal rather than for a specific kind of knowledge. 
One possible reason is that even though the designers of WSC are trying to avoid any statistical correlation between the answer and the trigger word, such statistical correlation still exists.
As a result, pre-trained language representation models can learn such correlation from large-scale training corpus and thus can answer WSC questions without fully understanding the reasons behind.
Besides that, another interesting finding is that GPT-2 (large), as the best unsupervised model on WSC, performs poorly on WinoWhy. 
One possible explanation is that a lot of negative examples are generated with GPT-2 (large), and thus the dataset brings extra challenges for GPT-2 (large).
Last but not least, we can find that fine-tuning over similar dataset (i.e., WSCR and WinoGrande) can slightly help RoBERTa, but the effect is still quite limited. This is probably because such a fine-tuning procedure only teaches pre-trained models to better answer WSC questions rather than understand the commonsense knowledge behind.

\subsection{Supervised Setting}

Besides the unsupervised setting, we are also interested in whether a model can learn to distinguish reasons through supervised learning. 

\subsubsection{Experiment Details}

\begin{table}[t]
\small
    \centering
    \begin{tabular}{c|c||c|c}
        \toprule
        Setting & Model & Accuracy & std\\
        \midrule
        \multirow{7}{*}{Five-fold (q)} & Glove + LSTM & 59.74\% &1.04\% \\
         & BERT (base) & 77.48\% &2.06\% \\
         & BERT (large) & 77.39\% &1.54\% \\
         & RoBERTa (base)& 75.01\% &2.48\% \\
         & RoBERTa (large)& 75.04\% &1.97\% \\
         & GPT-2 (small) & 74.48\% &2.43\% \\
         & GPT-2 (large) & 75.89\% &1.35\% \\
         \midrule
         \multirow{7}{*}{Five-fold (r)}& Glove + LSTM & 64.92\% &1.76\% \\
         & BERT (base)& 77.77\% &1.54\% \\
         & BERT (large)& 77.50\% &2.43\% \\
         & RoBERTa (base)& 74.41\% &1.35\% \\
         & RoBERTa (large)& 74.66\% &1.75\% \\
         & GPT-2 (small) &76.19\% &3.69\% \\
         & GPT-2 (large) &76.13\% &4.30\% \\
         \bottomrule
    \end{tabular}
    \caption{Accuracy and the standard deviation (std) results of evaluated supervised models.}
    \label{tab:winowhy-supervised}
    \vspace{-0.2in}
\end{table}

Here, we randomly divide the annotated dataset into five groups and conduct five-fold cross-validation. 
We tried two different splitting methods, one is based on the WSC questions and the other one is based on the reasons. 
We denote these two settings as Five-fold (q) and Five-fold (r) respectively. 
As WinoWhy can be viewed as a text classification task, we adopt the traditional encoding+classification framework and leverage a two-layer feed-forward neural network as the classification module. 
Seven different encoding methods (Bi-LSTM~\cite{hochreiter1997long}, BERT (base), BERT (large), GPT-2 (small), GPT-2 (large), RoBERTa (base), and RoBERTa (large)) are evaluated. 
For LSTM, we choose the number of layers to be two, the hidden embedding dimension to be 300, and Glove~\cite{DBLP:conf/emnlp/PenningtonSM14} to be the word embedding.
All models are trained for ten epochs.
Average accuracies over folds and standard deviations are reported.

\subsubsection{Result Analysis}

The results in Table~\ref{tab:winowhy-supervised} demonstrate that in general, WinoWhy is a challenging task as the best supervised model can only achieve 77.77\% accuracy on a two-class classification task.
Besides that, we also notice that all models are getting relatively large standard deviations, especially under the `Five-fold (r)' setting, which may imply that these supervised models are sensitive to the dataset distribution.
Both of these observations show that training a supervised model on WinoWhy is not enough to fully understand the reasons behind WSC decisions and we may need to include reasoning over more complex knowledge to solve this challenging problem.



\subsection{Discussion}

Based on the observations that fine-tuning over WSCR and WinoGrande can only help solve WSC rather than WinoWhy and the machine-learning based models over WinoWhy can be sensitive to the dataset distribution, it is reasonable to suspect that the improvement achieved by fine-tuning over a similar or same dataset might come from better dataset fitting rather than better commonsense reasoning. 
As the original purpose of proposing both WSC and WinoWhy is to evaluate how good current AI systems can understand commonsense knowledge rather than solve these questions by fitting the dataset, the unsupervised setting might be the more reasonable evaluation setting.


\section{Related Work}\label{sec:related-work}

As an important knowledge resource for many artificial intelligence systems, commonsense knowledge covers various knowledge categories like causality~\cite{sap2019atomic}, reasoning~\cite{schubert2015genuineunderstanding}, property~\cite{liu2004conceptnet}, and quantity~\cite{DBLP:conf/acl/ElazarMRBR19},
and has been proven crucial in many downstream tasks like question answering~\cite{lin2019kagnet}, dialogue system~\cite{zhou2018commonsense}, reading comprehension~\cite{wang2018yuanfudao}, and pronoun coreference resolution~\cite{levesque2012winograd}. 
Among all these tasks, Winograd Schema Challenge (WSC)~\cite{levesque2012winograd} is viewed as one of the most challenging ones because solving WSC questions typically requires inference over various kinds of commonsense knowledge.
Conventionally, people tried to solve WSC questions in an unsupervised way by leveraging either search engines~\cite{DBLP:conf/emnlp/EmamiCTSC18}, linguistic knowledge~\cite{DBLP:conf/acl/ZhangDS19,zhang2019aser}, or language representation models~\cite{DBLP:conf/acl/KocijanCCYL19}.
Experimental results showed that these models still cannot fully solve the problem but we are not clear about how to further improve them.
One important reason behind this is that the conventional definition of commonsense knowledge is too vague and thus we are not clear about what kinds of knowledge are still challenging for current commonsense reasoning models.
In this paper, we use the WSC task as the breaking point to conduct a deep diagnosis of essential commonsense knowledge types, which sheds some light on how to achieve a better commonsense reasoning system in the future.

\section{Conclusion}\label{sec:conclusion}

In this paper, we presented the first deep diagnosis of essential commonsense knowledge for answering Winograd Schema Challenge questions. 
By doing so, we better understand the strengths and limitations of current commonsense reasoning models. 
More importantly, we better know about what kinds of commonsense knowledge are required to be acquired for better commonsense reasoning.
On top of the collected reasons, we develop a new task called WinoWhy, which requires models to select the plausible reasons for answering WSC questions.
Experiments show that even though current models have gained significant improvement over the original WSC task, they still cannot fully understand the reasons behind.

\section*{Acknowledgements}

This paper was supported by the Early Career Scheme (ECS, No. 26206717) from the Research Grants Council in Hong Kong and the Tencent AI Lab Rhino-Bird Focused Research Program. 

\bibliography{WinoWhy}
\bibliographystyle{acl_natbib}

\clearpage

\appendix
\section{Reason Collection}\label{appendix:reason_collection}

\begin{figure*}[hbt]
    \centering
    \includegraphics[width=0.8\linewidth]{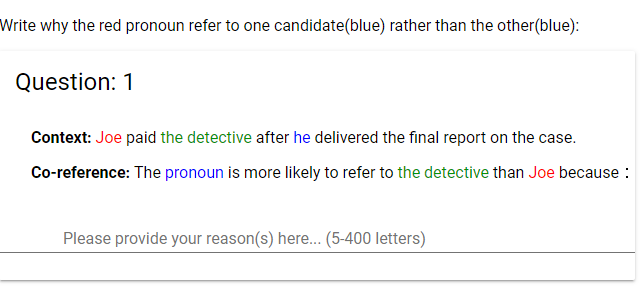}
    \caption{A screen shot of the reason collection survey.}
    \label{fig:appendix_collection_survey}
\end{figure*}

\begin{figure*}[hbt]
    \centering
    \includegraphics[width=0.8\linewidth]{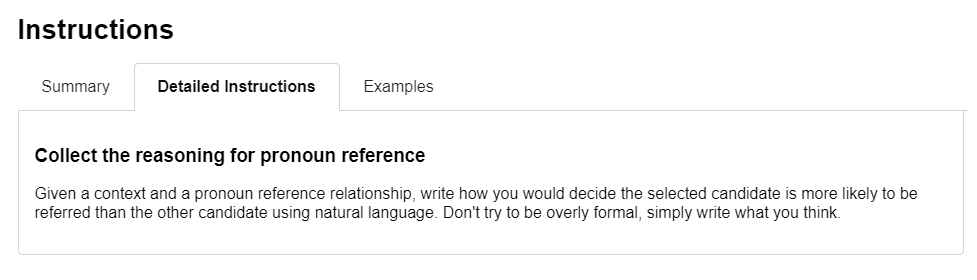}
    \caption{The instruction for the reason collection survey.}
    \label{fig:appendix_collection_instruction}
\end{figure*}

\begin{figure*}[hbt]
    \centering
    \includegraphics[width=0.8\linewidth]{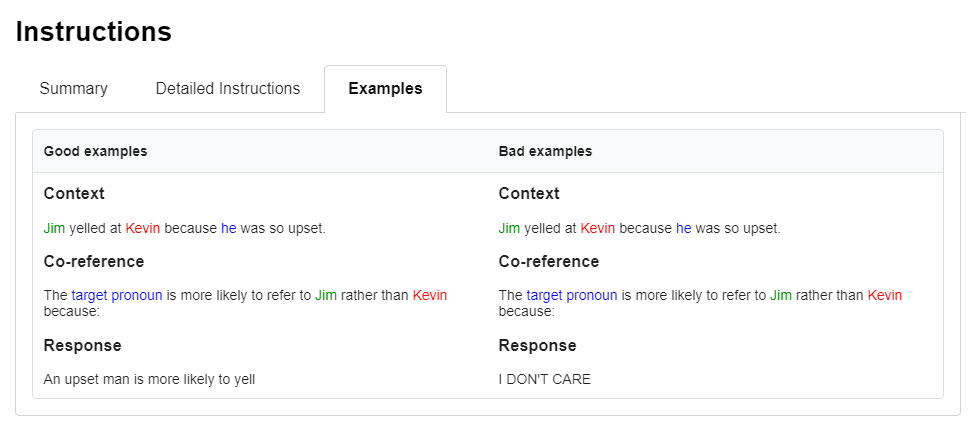}
    \caption{Examples for the reason collection survey.}
    \label{fig:appendix_collection_example}
\end{figure*}

\clearpage

\section{Reason Plausibility}\label{appendix:reason_plausibility}

\begin{figure*}[hb]
    \centering
    \includegraphics[width=0.8\linewidth]{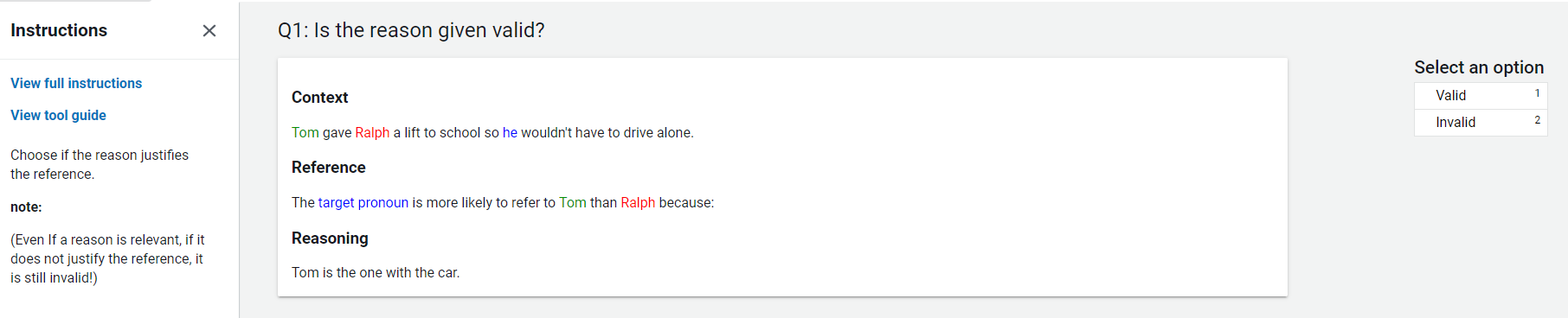}
    \caption{A screen shot of the plausibility annotation survey.}
    \label{fig:appendix_plausibility_survey}
\end{figure*}

\begin{figure*}[hb]
    \centering
    \includegraphics[width=0.8\linewidth]{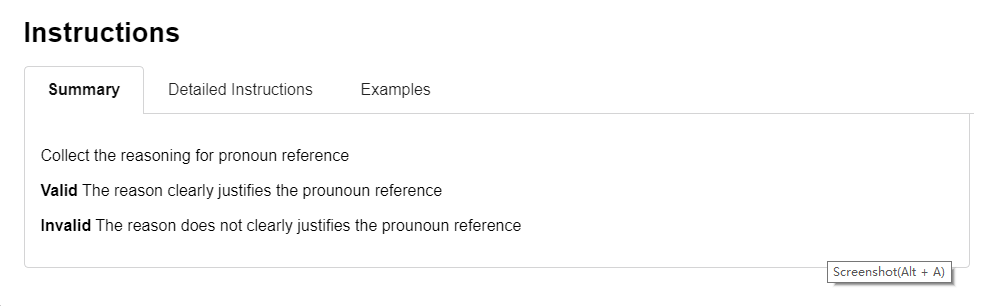}
    \caption{The first instruction for the plausibility annotation.}
    \label{fig:appendix_plausibility_instruction_1}
\end{figure*}

\begin{figure*}[hb]
    \centering
    \includegraphics[width=0.8\linewidth]{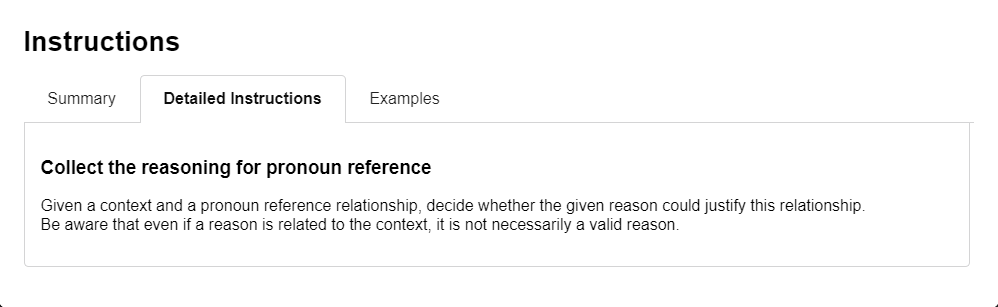}
    \caption{The second instruction for the plausibility annotation.}
    \label{fig:appendix_plausibility_instruction_2}
\end{figure*}

\begin{figure*}[hb]
    \centering
    \includegraphics[width=0.8\linewidth]{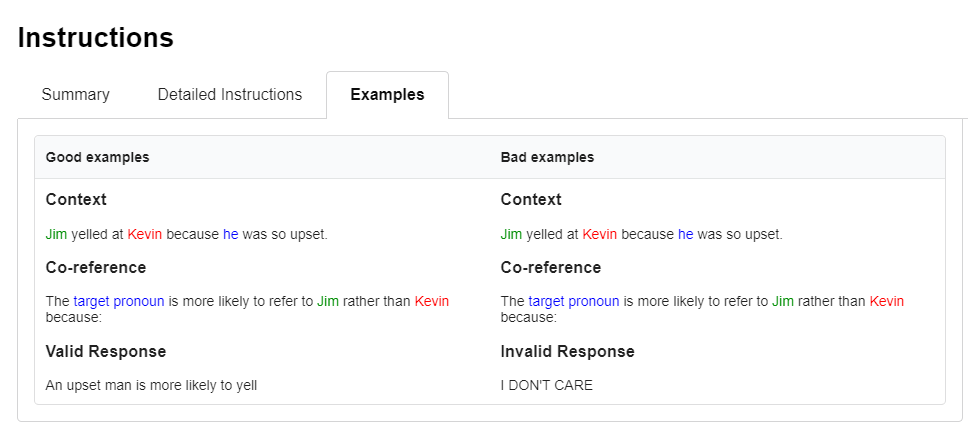}
    \caption{Examples for the plausibility annotation survey.}
    \label{fig:appendix_plausibility_example}
\end{figure*}

\clearpage
\section{Reason Categorization}\label{appendix:reason_categorization}

\begin{figure*}[hb]
    \centering
    \includegraphics[width=0.8\linewidth]{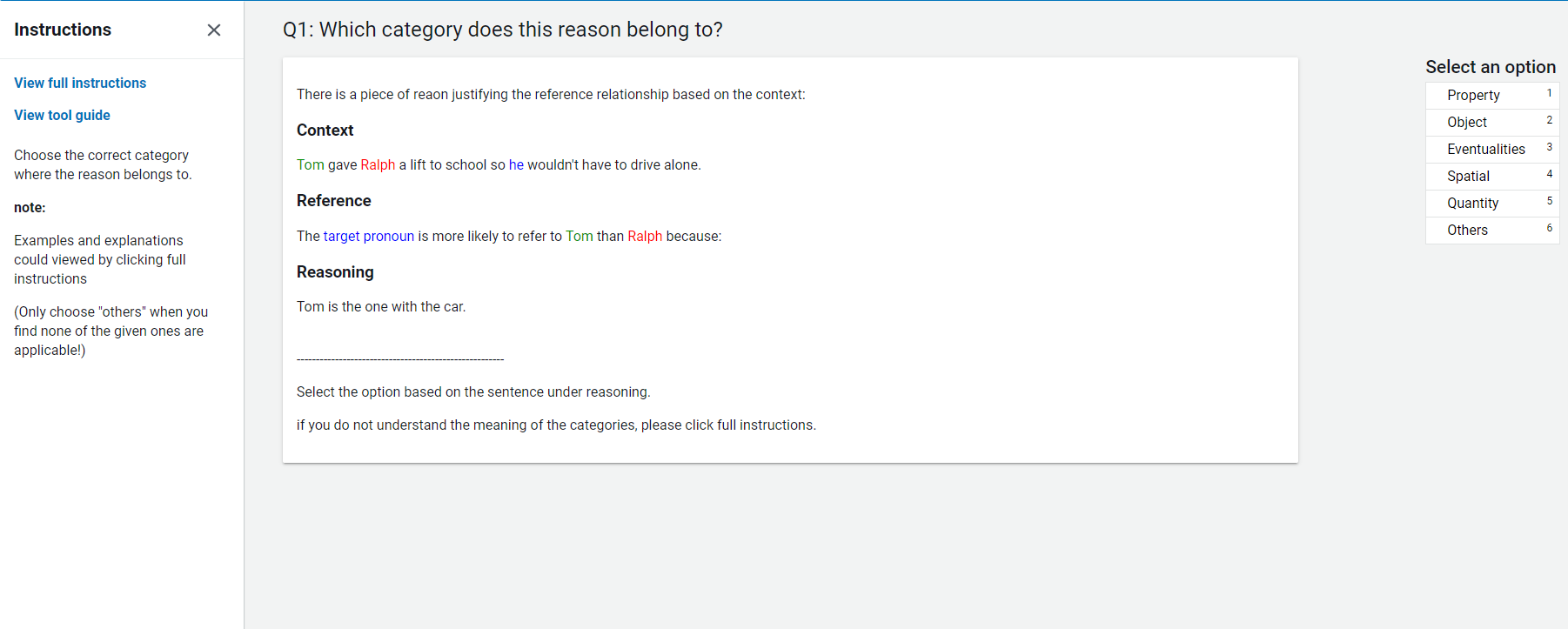}
    \caption{A screen shot of the reason categorization survey.}
    \label{fig:appendix_categorization_survey}
\end{figure*}

\begin{figure*}[hb]
    \centering
    \includegraphics[width=0.8\linewidth]{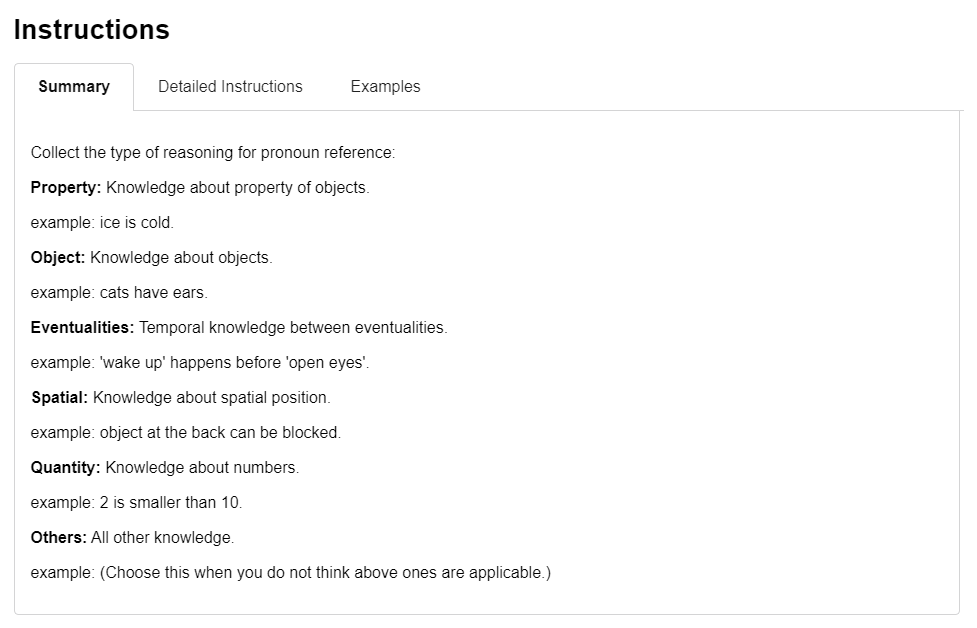}
    \caption{The first instruction for the reason categorization.}
    \label{fig:appendix_categorization_intruction_1}
\end{figure*}

\begin{figure*}[hb]
    \centering
    \includegraphics[width=0.8\linewidth]{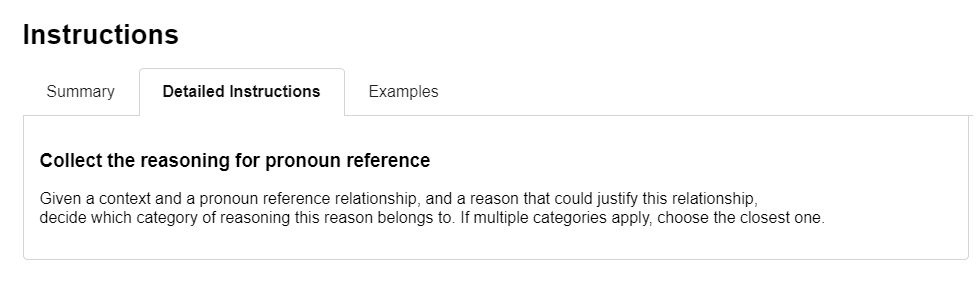}
    \caption{The second instruction for the reason categorization.}
    \label{fig:appendix_categorization_instruction_2}
\end{figure*}

\begin{figure*}[hb]
    \centering
    \includegraphics[width=0.8\linewidth]{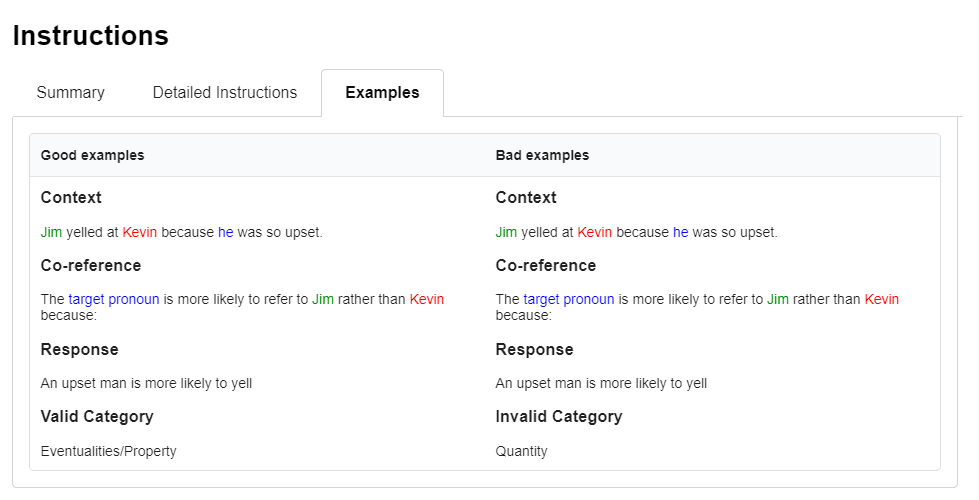}
    \caption{Examples for the reason categorization survey.}
    \label{fig:appendix_categorization_example}
\end{figure*}

\end{document}